\pdfoutput=1

\documentclass[11pt]{article}

\usepackage[preprint]{acl}

\usepackage{times}
\usepackage{latexsym}
\usepackage{booktabs}
\usepackage[T1]{fontenc}
\usepackage{multirow}

\usepackage[utf8]{inputenc}

\usepackage{microtype}

\usepackage{inconsolata}

\usepackage{graphicx}
\usepackage{algorithm}
\usepackage{algorithmic}
\usepackage{amsmath}
\usepackage{amssymb}
\usepackage{times}
\usepackage{soul}
\usepackage{url}
\usepackage{amsmath}
\usepackage{amsthm}
\usepackage{booktabs}
\usepackage[switch]{lineno}
\usepackage{colortbl}
\usepackage{multirow}
\title{
RATE-Nav: Region-Aware Termination Enhancement \\ 
for Zero-shot Object Navigation with Vision-Language Models}

\author{
 \textbf{Junjie Li\textsuperscript{1,2}},
 \textbf{Nan Zhang\textsuperscript{2}},
 \textbf{Xiaoyang Qu\textsuperscript{2}},
 \textbf{Kai Lu\textsuperscript{1}},
 \textbf{Guokuan Li\textsuperscript{1,\protect\footnotemark[2]}},
 \textbf{Jiguang Wan\textsuperscript{1}} and
 \textbf{Jianzong Wang\textsuperscript{2,\protect\footnotemark[2]}}
 \\
  \textsuperscript{1}Huazhong University of Science and Technology, Wuhan, China,
\\ 
\textsuperscript{2}Ping An Technology (Shenzhen) Co., Ltd., Shenzhen, China
\\
 \small{
   \textbf{Correspondence:} \href{mailto:liguokuan@hust.edu.cn}{liguokuan@hust.edu.cn} and \href{mailto:jzwang@188.com}{jzwang@188.com}
}
}

\begin{document}
\maketitle
\footnotetext[2]{Jianzong Wang and Guokuan Li are joint corresponding authors.}
\begin{abstract}
Object Navigation (ObjectNav) is a fundamental task in embodied artificial intelligence. Although significant progress has been made in semantic map construction and target direction prediction in current research, redundant exploration and exploration failures remain inevitable. A critical but underexplored direction is the timely termination of exploration to overcome these challenges. We observe a diminishing marginal effect between exploration steps and exploration rates and analyze the cost-benefit relationship of exploration. Inspired by this, we propose RATE-Nav, a Region-Aware Termination-Enhanced method. It includes a geometric predictive region segmentation algorithm and region-Based exploration estimation algorithm for exploration rate calculation. By leveraging the visual question answering capabilities of visual language models (VLMs) and exploration rates enables efficient termination.RATE-Nav achieves a success rate of 67.8\% and an SPL of 31.3\% on the HM3D dataset. And on the more challenging MP3D dataset, RATE-Nav shows approximately 10\% improvement over previous zero-shot methods.

\end{abstract}

\section{Introduction}
Object navigation \cite{dang2023search,campari2020exploiting}, as one of the core capabilities of embodied agents, aims to enable agents to autonomously locate and navigate to specified target objects in unknown environments \cite{batra2020objectnav,sun2024survey}. This capability is crucial for practical applications such as service robots, for instance, in performing fetch tasks in home environments. However, in real-world applications, robots often operate in unknown and dynamically changing environments, and search targets are not limited to predefined categories, posing significant challenges to the zero-shot generalization ability of navigation systems\cite{majumdar2022zson}. 

\begin{figure}[!t]
\centering
\includegraphics[width=0.99\columnwidth]{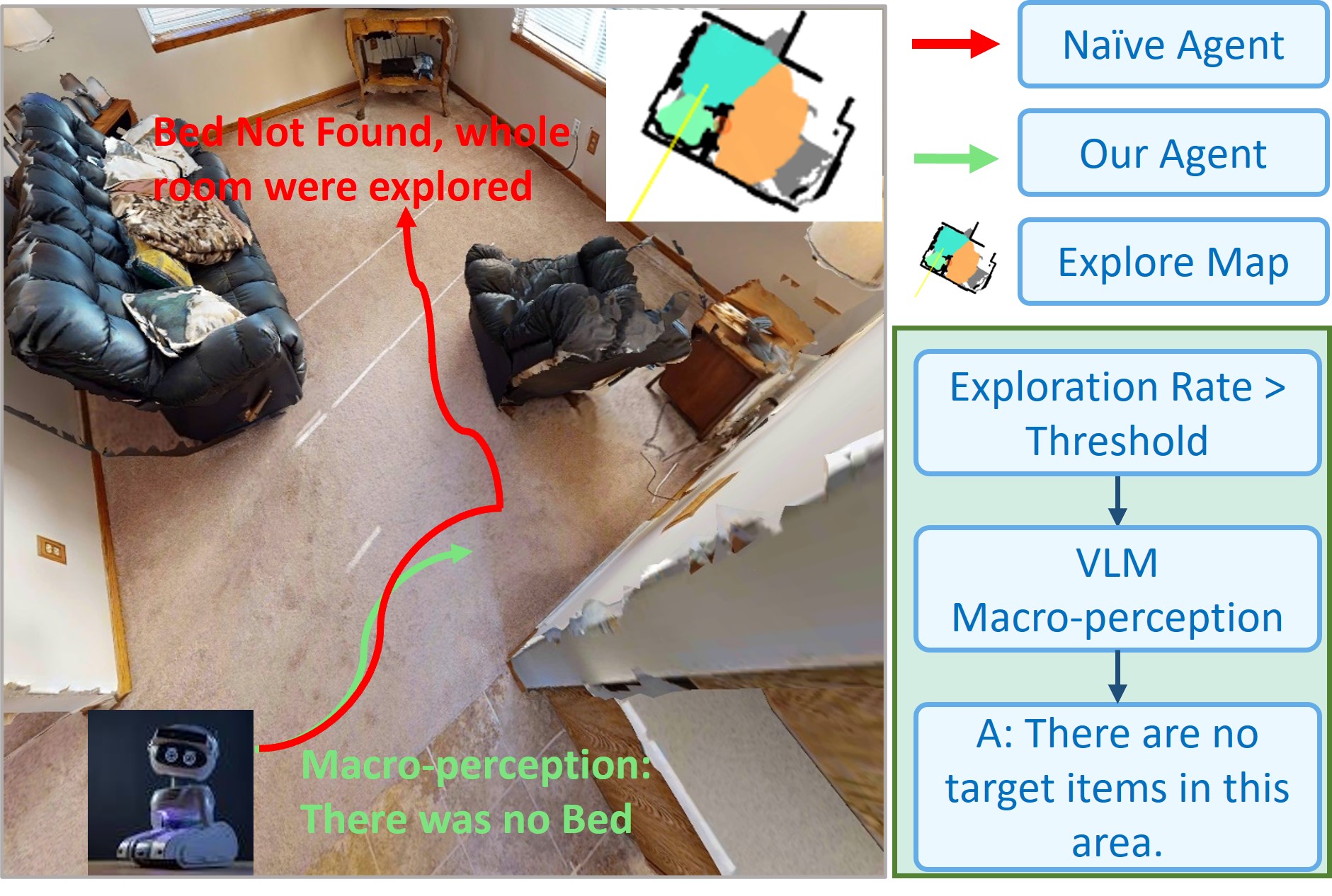} 
\caption{The main difference between our method and existing approaches is the exploration strategy. Naive agent fully explores the current region before moving to the next. In contrast, our agent uses the commonsense reasoning of VLMs to decide whether to continue exploring the current region after achieving a certain exploration level. 
}
\label{fig1}
\end{figure}



With the rapid development of Large Language Models (LLMs) \cite{ achiam2023gpt} and Vision-Language Models (VLMs) \cite{ radford2021learning,li2023blip,liu2024llava}) with powerful reasoning capabilities in the field of artificial intelligence, 
researchers \cite{zhou2023esc, shah2023navigation} have turned to employing these models for intermediate goal planning: first, by modeling environmental observations with VLMs to construct semantically rich scene representations, and then utilizing the commonsense reasoning abilities embedded in LLMs to predict the possible locations of target objects.
Methods \cite{zhou2023esc, yin2024sgnav} based on LLMs and VLMs have demonstrated notable advantages navigation strategies, thereby enhancing the performance of navigation systems to some extent.

Existing research relies on comprehensive exploration, but there remains a research gap in the design of exploration termination strategies. 
As shown in \textbf{Figure \ref{fig1}}, traditional navigation strategies struggle to accurately assess the current exploration state when searching specific areas, typically requiring a complete search of the current area before moving to a new one. 
From a marginal utility perspective, this strategy is inefficient: in the early stages of information gathering, the investment yields significant returns, but as exploration deepens, the marginal value of new information gained from each operation gradually diminishes.This prompts us to consider whether the search in the current area could be terminated earlier. To validate this hypothesis, we conducted hundreds of navigation experiments on the HM3D dataset, thoroughly analyzing the relationship between exploration cost and benefit. Detailed analysis can be found in  \textbf{Chapter \ref{sec:marginal}}.

Furthermore, existing research primarily relies on exploration maps constructed by visual perception to identify target-free regions. Precision errors and model limitations make it difficult to fully annotate these regions on the map. 
As shown in Figure \ref{overview}, a situation may occur where a region is largely explored, but repeated boundary settings are triggered due to a small unknown area. This results in redundant exploration, which not only wastes resources but also reduces the adaptability and flexibility of the navigation system. This inspires us to consider whether we can estimate explored region to avoid repeated exploration and thereby improve efficiency.

Based on our analysis, we propose \textbf{RATE-Nav} (Region-Aware Termination-Enhanced Navigation), a novel navigation method that enhances exploration efficiency through region-level search and intelligent termination strategies. As one of its key components, we develop a Geometric Predictive Region Segmentation module that assists in partitioning incomplete environmental maps into relatively independent regions. This segmentation module utilizes geometric features to predict unexplored areas, contributing to the transformation from point-by-point search into a region-level search problem. The region-based approach enables holistic evaluation of environmental segments, leading to better estimation of unknown spaces and improved navigation efficiency. For the critical task of region-level evaluation, we leverage VLMs which excel at macroscopic environmental perception. When VLMs determine that a target object is unlikely to exist within a region, the system designates it as an "exploration-free zone," immediately terminating current exploration and preventing future redundant searches in that area. This VLM-based termination strategy effectively reduces unnecessary exploration and substantially improves overall efficiency.

Our contributions are summarized as follows:

\begin{itemize}
    \item The marginal utility between navigation efficiency and information acquisition was investigated, which reveals the information gained per step decreases as exploration progresses, suggesting that comprehensive exploration is not always necessary.
    \item  A novel zero-shot object navigation method called RATE-Nav inspired by marginal utility was proposed, which uses VLMs macro-perception and \textbf{R}egion \textbf{A}ware to determine exploration \textbf{T}ermination, thereby achieving effective navigation \textbf{E}nhancement.
    
    \item Extensive experiments are conducted to demonstrate that RATE-Nav significantly outperforms existing zero-shot object navigation methods.
\end{itemize}

\section{Related Works}
\subsection{Zero-shot Object Navigation}
Research in Object Navigation \cite{li2022object} has evolved from early end-to-end \cite{khandelwal2022simple} deep learning approaches (based on RL \cite{Dang_2023_ICCV} and IL \cite{Ramrakhya_2023_CVPR}) to modular approaches using dynamic map representations \cite{NEURIPS2020_2c75cf26, Zhang_2023_CVPR, ramakrishnan2022poni, chen2023not} to improve computational efficiency and adaptability. With the advancement of large models and increasing demands for generalization in object navigation, researchers have proposed various unsupervised \cite{majumdar2022zson} and zero-shot \cite{gadre2023cows} methods. For instance, zero-shot approaches \cite{gadre2023cows, TriHelper} enable agents to navigate to unseen target categories by leveraging general knowledge from training, while ESC \cite{zhou2023esc} and OpenFMNav \cite{kuang2024openfmnav} emphasize the importance of commonsense reasoning and VLM perception. VorNav \cite{wu2024voronav} demonstrated the potential of novel map representations.

\subsection{Large Pre-trained Models for Zero-shot Object Navigation}
Large pre-trained models \cite{achiam2023gpt, Qwen-VL, naveed2023comprehensive} have revolutionized object navigation by enabling zero-shot decision-making in unknown environments, with works leveraging VLMs for scene understanding \cite{zhou2023esc, kuang2024openfmnav}, LLMs for object-room relationship analysis \cite{yu2023l3mvn, cai2024bridging}, and advanced prompting techniques for navigation planning \cite{long2024instructnav, yin2024sgnav}. Building upon these advances, we propose to address an even more challenging scenario of zero-shot object navigation without any navigation training data.

\section{Motivation}
\subsection{Marginal Utility in Object Navigation}
\label{sec:marginal}
The concept of marginal utility in economics is used to explain the diminishing returns as consumption increases. We observe a similar relationship between exploration steps and exploration rate in Object Navigation.

\begin{figure}[!th]
\centering
\includegraphics[width=0.99\columnwidth]{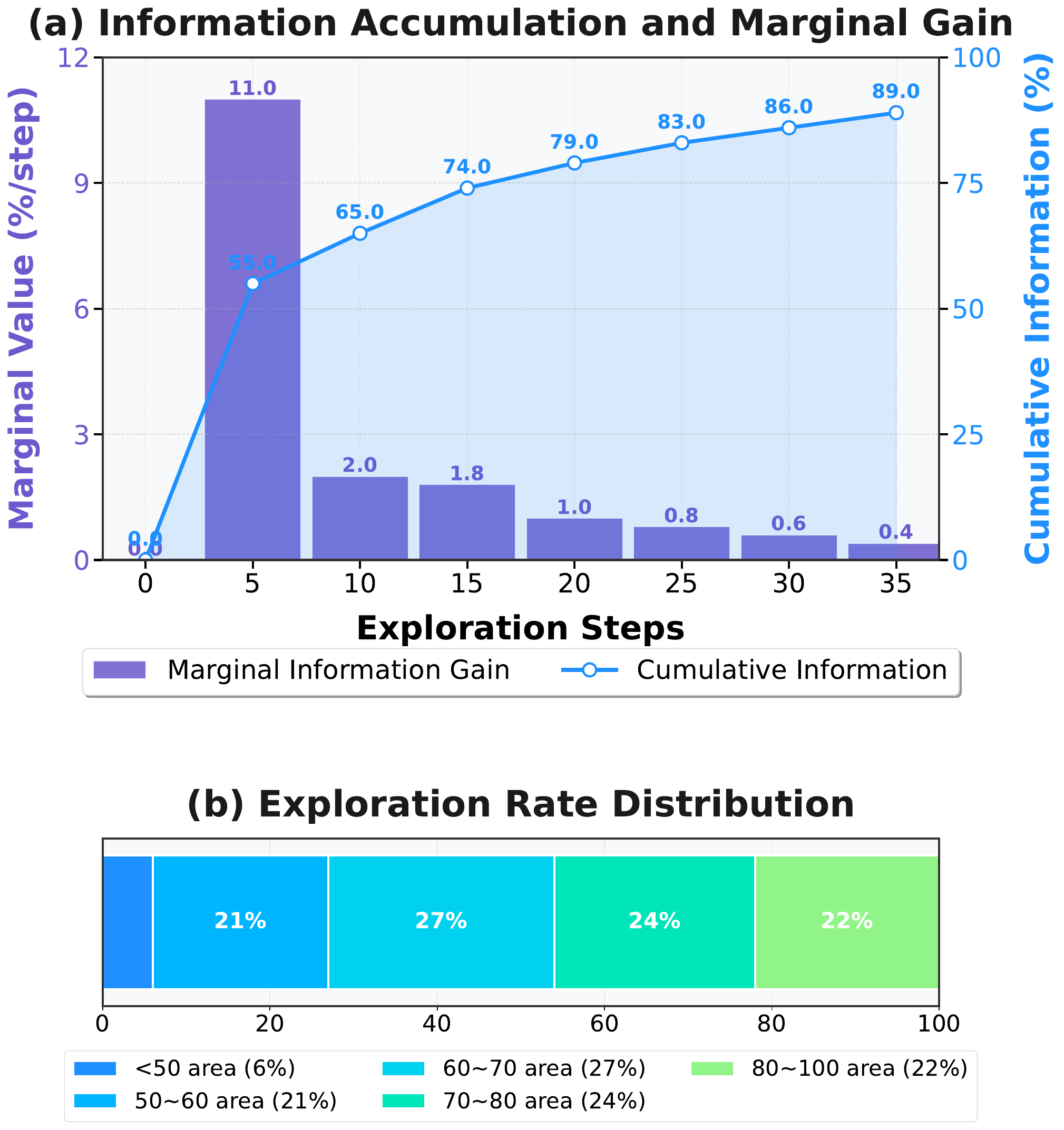} 
\caption{
Systematic analysis of navigation efficiency and information acquisition.In (a), the x-axis is exploration steps, and the line graph shows cumulative information. By step 5, exploration reaches 55\% of the region. The bar chart displays marginal values (e.g., $55\% \div 5 = 11\%/step$ for steps 0-5). In (b), 6\% of samples find the target after exploring 50\% of the region, increasing to 78\% at 80\% exploration.
}
\label{fig2}
\end{figure}

To investigate the planning allocation problem in navigation, we conducted a systematic analysis of navigation efficiency and information acquisition using the HM3D dataset. Our study examined the relationship between exploration steps and exploration rate, as illustrated in \textbf{Figure \ref{fig2}}. The analysis comprises two key components: a) evaluating the correlation between exploration cost and coverage rate during area exploration, and b) analyzing the distribution of exploration coverage rates in regions where robots successfully locate targets. As shown in \textbf{Figure \ref{fig2}a}, our findings reveal a significant diminishing return in new environmental information gained per step as exploration continues. Notably, this exhaustive exploration approach is not always necessary. The results from \textbf{Figure \ref{fig2}b} indicate that the detection of targets occurs primarily within two distinct phases: the Efficient Acquisition Phase and the Stable Exploration Phase.

Based on the statistical results, we divide the area exploration process into three characteristic phases. First is the Efficient Acquisition Phase, where robots can quickly gather environmental information with minimal movement due to their field of view advantage. The second is the stable exploration phase, during which robots continuously move to steadily acquire new environmental information. Finally, there is the Edge Completion Phase, where robots need to employ complex strategies such as detouring to gather remaining information due to the presence of obstacles.



Through the above research, we introduce the concept of marginal cognition, which identifies two key issues in exploration: 1) Core information collected in the early stages typically holds the highest value, while subsequent marginal information yields diminishing returns; 2) Initial information gathering efforts produce significant gains, but the knowledge acquired per unit time (or cost) tends to decrease as exploration deepens.

\subsection{Challenges and Inspiration}
\label{sec:challenges}
While recent studies have attempted to enhance target selection in foundation models through richer environmental information integration, redundant exploration and exploration failures remain inevitable. These challenges underscore the importance of making rapid decisions based on partial exploration data, avoiding the collection of lower-value information at higher costs, and efficiently backtracking from unproductive paths to improve navigation efficiency. 

The marginal utility in object navigation inspires us to explore whether marginal effects can help an agent dynamically adjust its behavior based on changes in the environment. Specifically, in object navigation, the agent could perform region-aware termination based on marginal effects. That is, when the marginal effect of exploring a particular region becomes sufficiently low, the agent may consider halting further exploration in that region and redirecting its efforts to more promising areas.


\section{Methodology}

\subsection{Method Overview}
\begin{figure*}[t]
\centering
  \includegraphics[width=0.99\linewidth]{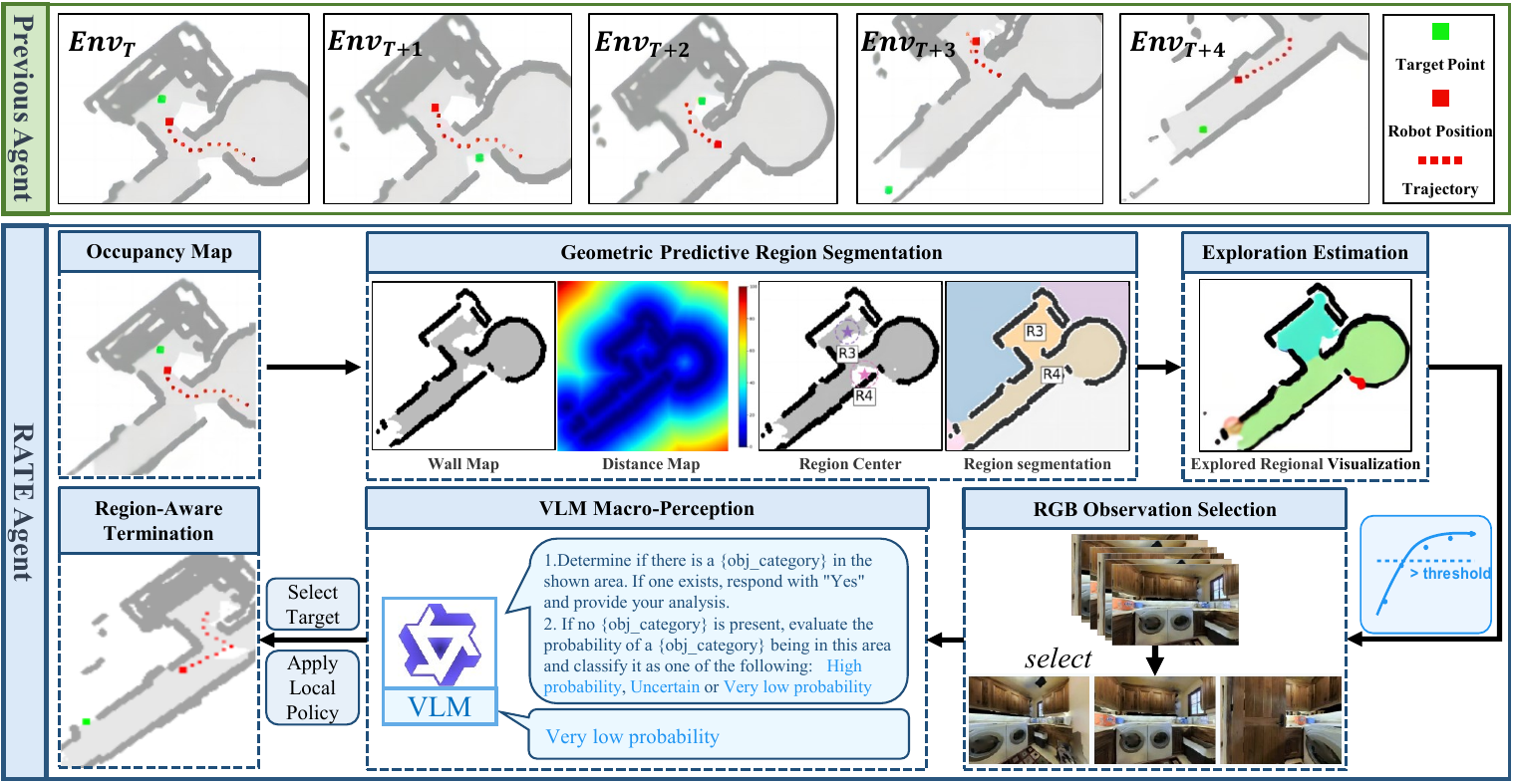}
  \caption {The workflow of RATE-Nav.The previous agent's requirement for complete spatial exploration often leads to redundant exploration efforts, as evidenced by the repeated exploration of identical regions at $Env_T$ and $Env_{T+2}$. In contrast, the RATE agent effectively minimizes redundant exploration in low-value areas.}
    \label{overview}
\end{figure*}

To address the challenges mentioned in \textbf{Chapter \ref{sec:challenges}}, we propose a macro-perception strategy that leverages VLMs and region exploration rate to evaluate the potential of continuous exploration, enabling efficient exploration termination. The RATE-Nav workflow is illustrated in \textbf{Figure \ref{overview}} and \textbf{Algorithm \ref{alg:region_exploration}}. Our Geometric Predictive Region Segmentation algorithm utilizes geometric features to predict and segment unexplored areas within the map, providing a robust solution to the challenge of ambiguous region references in VLM outputs. This geometric feature-based approach enables more accurate and meaningful region partitioning. Following region segmentation, we introduce a method for estimating explored area that considers both the robot's visible areas and traversable spaces. When the marginal utility of exploring a specific region drops below a threshold (indicated by the region exploration rate), we filter key frame inputs from RGB observations and feed them into the VLM for macro-environmental perception. The VLM's logical reasoning capabilities then determine whether to terminate further exploration of that region, preventing redundant exploration in low-value areas.

\subsection{Region Semantic Map}
The Semantic Map serves as a crucial information tool for evaluating candidate points, playing a dual role in this research: Firstly, it utilizes its rich language information to predict potential locations of target objects based on common sense reasoning. Besides, it builds customized prompts based on identified object features in each Region, enhancing VLMs accuracy in assessing area candidacy.


To construct a high-quality Semantic Map, we employ ConceptGraphs \cite{gu2024conceptgraphs} for environment modeling. The specific process includes extracting semantic features from RGB-D sequences, accurately projecting them onto 3D point clouds, and optimizing through multi-view fusion, ultimately generating a complete dataset containing 3D object information and their visual and language descriptors. By combining this information with spatial position data, we can obtain comprehensive semantic information about objects contained in each area, providing reliable prior knowledge support for subsequent target object localization.

\subsection{Geometric Predictive Region Segmentation}
To characterize explored areas and define region boundaries, map segmentation is essential. We segment the map based on tall obstacles (primarily walls) that obstruct robot vision into distinct regions. While these regions often correspond to different rooms from an indoor navigation perspective, larger spaces like living rooms may be divided into multiple regions to facilitate description and computation. Given RGB-D images ${I_0, ..., I_t}$ and robot poses ${p_0, ..., p_t}$, we use 3D point cloud modeling with height threshold $h$ to identify vision-obstructing obstacles, collectively termed as the wall map $W$.
Based on the wall map, we propose a watershed algorithm-based region segmentation method, as shown in \textbf{Algorithm \ref{alg:region_exploration} Phase 1}. The method consists of these steps:

1) Wall preprocessing: Apply distance transform $D_w$ to the wall map and mark areas within a threshold $\delta$ (1.5 units) of walls as wall regions.

2) Distance map generation: Perform Euclidean distance transform $D_e$ on the processed binary map to create a distance map.

3) Region center detection: Identify potential region centers $C$, which contains center points from $c_1$ to $c_n$, using local maxima detection on the distance map. A point $c_i$ (with coordinates $(x,y)$) is considered a center if its value $D_e(x,y)$ on the distance map exceeds a threshold $\tau$ and is maximal within its neighborhood $N(x,y)$. In other words, $D_e(x,y)$ must be equal to the maximum $D_e$ value among all points $(x',y')$ within its neighborhood $N(x,y)$.

4) Region segmentation: Apply watershed algorithm to expand detected region centers, using each center as a seed point $s_i$ and simulating a "flooding" process for automatic segmentation. The region label $R(x,y)$ is assigned as:
\begin{equation}
R(x,y) = \arg\min_{i} \{P(x,y,s_i)\}
\end{equation}
where $P(x,y,s_i)$ represents the topographical distance from point $(x,y)$ to seed point $s_i$.

5) Post-processing optimization: Refine initial segmentation by merging small regions below a threshold $\alpha$ with adjacent larger regions:
\begin{equation}
\small
R'(x,y) = \begin{cases}
\arg\max_{k\in N_R(i)} |R_k| & \begin{array}{l}
\text{if } |R_i| < \alpha
\end{array} \\
R(x,y) & \text{otherwise}
\end{cases}
\end{equation}
where $|R_i|$ represents the size of region $i$ and $N_R(i)$ represents the set of regions adjacent to region $i$.

This method produces semantically meaningful region segmentation with clear boundaries and rich geometric feature descriptions. The approach considers both physical connectivity and practical utility through post-processing, establishing a solid spatial representation foundation for subsequent exploration and navigation tasks.





\begin{algorithm}[t]
\caption{Region Exploration Algorithm}
\label{alg:region_exploration}
\begin{algorithmic}[0]
\STATE \textbf{Input:} RGB-D images $\mathcal{I} = \{I_0...I_t\}$
\STATE Robot poses $\mathcal{P} = \{p_0,...,p_t\}$
\STATE Target category $c$
\end{algorithmic}
\begin{algorithmic}[1]
\STATE \textbf{Phase 1: Region Map Construction}
\STATE $S \leftarrow \text{BuildSemanticMap}(\mathcal{I})$
\STATE $R \leftarrow \text{SegmentRegions}(S)$

\STATE \textbf{Phase 2: Exploration Estimation}
\STATE $V \leftarrow \text{CalculateVisibleArea}(\mathcal{P})$
\STATE $E \leftarrow \text{EstimateRegionRate}(V, R)$

\STATE \textbf{Phase 3: VLM Assessment}
\STATE $F \leftarrow \text{SelectKeyFrames}(\mathcal{I}, R, E)$
\STATE $P \leftarrow \text{VLMAssessment}(F, c)$

\STATE \textbf{Phase 4: Decision Making}
\IF{$\text{IsLowRelevance}(P)$}
    \STATE $R.\text{priority} \leftarrow \text{low}$
\ELSE
    \STATE $\text{ContinueSearch}(R, F)$
\ENDIF
\end{algorithmic}
\end{algorithm}




\subsection{Region-Based Exploration Estimation}
After obtaining the region segmentation map, we need to estimate the robot's visible areas to determine the coordinates of explored regions, as presented in \textbf{Algorithm \ref{alg:region_exploration} Phase 2}. Given the robot's positions $loc_0, \ldots, loc_t$ and orientation information $d_0, \ldots, d_t$, with a maximum visible distance $d_{max}$ defined. Based on the wall map, we calculate the visible area using the following formula:
\begin{equation}
\small
V_t = \{p \mid \|p - loc_t\| \leq d_{max} \&
\text{LoS}(loc_t, p) = \text{True}\}
\end{equation}

where $\text{LoS}(loc_t, p)$ is the line-of-sight function that returns True if there is an unobstructed path between points $loc_t$ and $p$, implemented using ray tracing Bresenham algorithms.

Additionally, combining with existing mapping methods, we can obtain the traversable area $M$ and obstacle point set $O$. By taking their intersection, we can get the estimated exploration area point set $E$, as shown in the formula:
\begin{equation}
E = \bigcup_{t=0}^T (V_t \cup M_t)
\end{equation}
This equation calculates the total explored area by taking the union of visible areas and traversable areas at each time step ($(V_t \cup M_t)$), then taking the union of these combined areas over the entire time period $T$.

After obtaining the exploration area point set, we can calculate the exploration rate $r$ for each region based on the above region segmentation:
\begin{equation}
r = \frac{|E \cap R_i|}{|R_i|}
\end{equation}
where $R_i$ represents the point set of the $i$-th region. The exploration rate $r$ indicates the proportion of the region that has been explored, calculated as the ratio of explored points to total points in the region.


\subsection{VLMs Macro-Perception of Termination Enhancement}
By setting an exploration threshold, when the area's exploration rate exceeds this threshold, the system triggers VLM for macro-environmental perception. During exploration, the system retains K key frames and records the visible range corresponding to each frame. When initiating VLM environmental macro-perception, the system intelligently filters these K frames based on two core criteria: 
1) Field-of-view priority: Select images whose field of view primarily covers the current target area
2) Exploration contribution: Selected images must significantly contribute to the overall exploration rate of the area

The filtered image set is then input to VLM for analysis. Through carefully designed prompts, VLM is guided to make a three-level probability assessment of target object presence in the current area: High probability, Uncertain, and Very low probabilityas. When VLM outputs Very low probability, the system assigns a very low exploration priority score to that area, avoiding redundant exploration in low-value areas and thereby improving overall exploration efficiency, which corresponds to \textbf{Phase 3 and 4} in \textbf{Algorithm \ref{alg:region_exploration} }.

After obtaining region assessments, we score candidate points by combining the region semantic map with Frontier-based Exploration (FBE), and generate specific actions through local policies such as the Fast Marching Method (FMM) \cite{sethian1999fast} once target points are determined. Furthermore, leveraging the powerful capabilities of VLMs, we perform re-perception of discovered targets to enhance detection accuracy.

\section{Experiment}

\begin{table*}[ht]
\centering
\caption{Comparison with previous work on MP3D and HM3D.}
\renewcommand{\arraystretch}{1.2}
\begin{tabular}{lcccccccccc}
\hline
\multirow{2}{*}{Method} & \multirow{2}{*}{Unsupervised} & \multirow{2}{*}{Zero-shot} & \multicolumn{2}{c}{MP3D} & \multicolumn{2}{c}{HM3D}  \\
\cline{4-7}
& & & SR $\uparrow$ & SPL $\uparrow$ & SR $\uparrow$ & SPL $\uparrow$ \\
\hline
SemEXP \cite{chaplot2020object} & No & No & 36.0 & 14.4 & - & -  \\
PONI \cite{ramakrishnan2022poni} & No & No & 31.8 & 12.1 & - & - \\
\hline
ZSON \cite{majumdar2022zson} & Yes & No & 15.3 & 4.8 & 25.5 & 12.6 \\
\hline
CoW \cite{gadre2023cows} & Yes & Yes & 7.4 & 3.7 & - & -  \\
TriHelper \cite{TriHelper} & Yes & Yes & - & - & 56.5 & 25.3 \\
ImagineNav \cite{zhao2024imaginenav} & Yes & Yes & - & - & 53.0 & 23.8 \\
ESC \cite{zhou2023esc} & Yes & Yes & 28.7 & 14.2 & 39.2 & 22.3  \\
L3MVN \cite{yu2023l3mvn} & Yes & Yes & 34.9 & 14.5 & 48.7 & 23.0  \\
VLFM \cite{yokoyama2024vlfm} & Yes & Yes & 36.2 & 15.9 & 52.4 & 30.3 \\
OpenFMNav \cite{kuang2024openfmnav} & Yes & Yes & 37.2 & 15.7 & 52.5 & 24.1  \\
ImagineNav-Oracle \cite{zhao2024imaginenav} & Yes & Yes & - & - & \underline{62.1} & \underline{31.1} \\
SG-Nav \cite{yin2024sgnav} & Yes & Yes & \underline{40.2} & \underline{16.1} & 54.2 & 24.1 \\

\cellcolor{gray!20}RATE-Nav & \cellcolor{gray!20}Yes & \cellcolor{gray!20}Yes & \cellcolor{gray!20}\textbf{50.3} & \cellcolor{gray!20}\textbf{20.6} & \cellcolor{gray!20}\textbf{67.8} & \cellcolor{gray!20}\textbf{31.3}\\
\hline
\end{tabular}
\end{table*}

\subsection{Experimental Setup and Implementation Details}
\textbf{Dataset}: We evaluate the effectiveness and navigation efficiency of our proposed method on two widely-used ObjectNav datasets: HM3D \cite{ramakrishnan2021habitat} and MP3D \cite{chang2017matterport3d} in the Habitat simulator. The HM3D validation dataset comprises 20 high-fidelity reconstructions of entire buildings and 2K validation episodes for object navigation tasks across six goal object categories. The MP3D validation dataset contains 11 indoor scenes, 21 object goal categories, and 2,195 object-goal navigation episodes.

\textbf{Evaluation Metrics}: We use three metrics to evaluate algorithm performance: 1) Success Rate (SR): percentage of successful episodes; 2) Success weighted by Path Length (SPL): combines success rate and path efficiency; 3) Soft SPL (SSPL): enhanced version of SPL providing finer evaluation by considering final agent-target distance. Higher values indicate better performance for all metrics.

\textbf{Implementation Details}: Experimental setup: max 500 steps per episode, agent camera at 0.88m height with 79° HFOV, discrete actions (0.25m forward step, 30° rotation). Using YOLO-World and GLIP for object detection with 640×640 RGB-D images. Agent maintains 800×800 2D occupancy map (0.05m/cell). Qwen-vl-max \cite{Qwen-VL} for complex perception, quantized Llama-Vision \cite{touvron2023llama} 11B for simple reasoning.

\subsection{Comparison With Prior Work}
\textbf{Baselines}: Based on supervision requirements and zero-shot capabilities, we categorize existing approaches into three groups. For non-zero-shot methods, we selected both supervised approaches (e.g., SemEXP \cite{chaplot2020object}, PONI \cite{ramakrishnan2022poni}) and unsupervised approaches (e.g., ZSON \cite{majumdar2022zson}) for comparison. In the zero-shot category, where most methods are unsupervised, we included several state-of-the-art approaches for comprehensive comparison, including OpenFMNav \cite{kuang2024openfmnav}, ImagineNav \cite{zhao2024imaginenav}, and SG-Nav \cite{yin2024sgnav}.

Our experimental results demonstrate that RATE-Nav significantly outperforms existing object navigation methods across supervised, unsupervised, and zero-shot categories, as shown in Table 1. On the HM3D dataset, RATE-Nav achieves a success rate of 67.8\% and an SPL of 31.3\%, substantially surpassing other methods. On the more challenging MP3D dataset, our approach shows approximately 10\% improvement over previous zero-shot methods. Notably, the significant improvement in SPL validates that our strategy of transforming point-to-point navigation into region-to-region navigation effectively enhances the robot's navigation efficiency.

\begin{figure}[h]
\centering
  \includegraphics[width=0.98\linewidth]{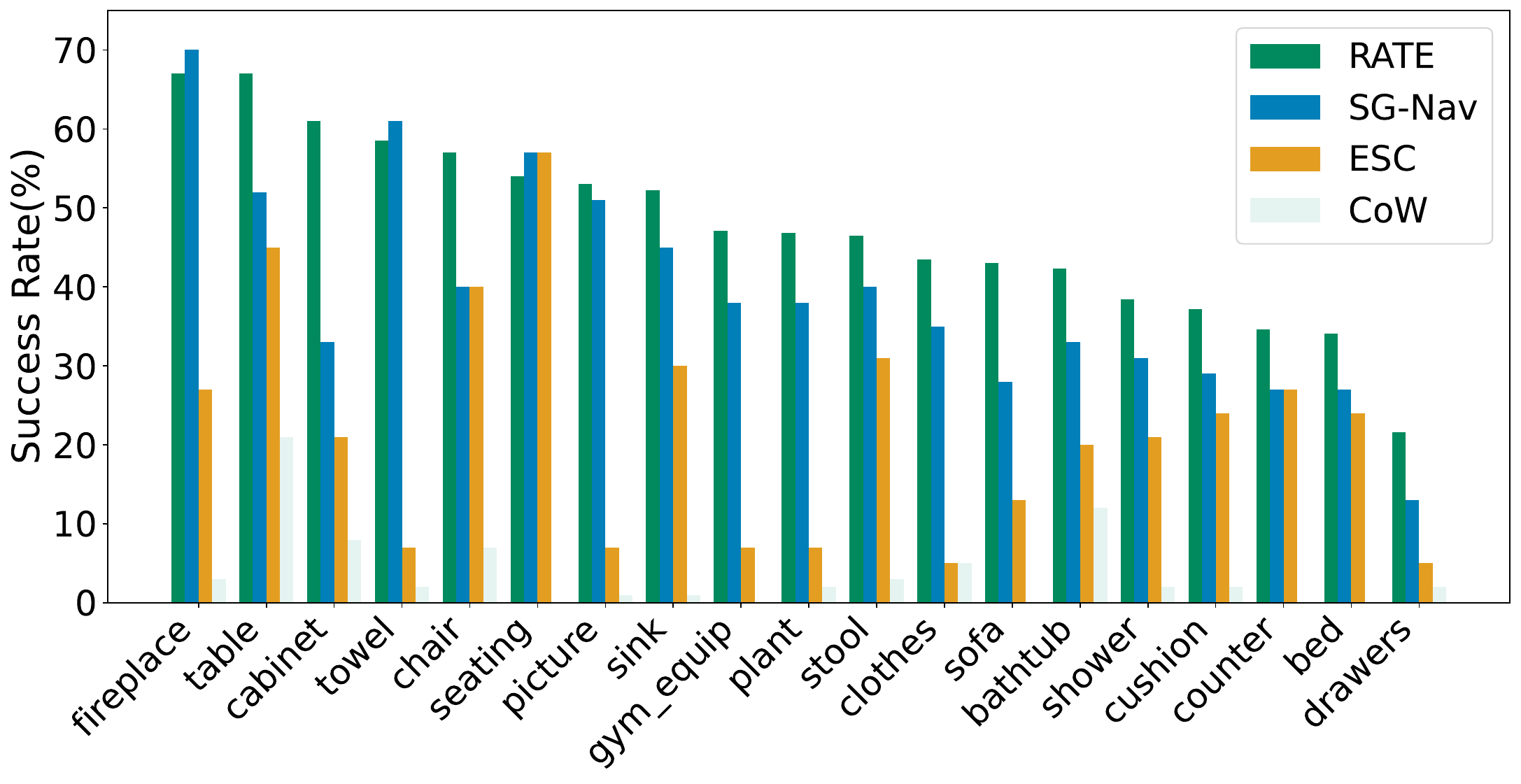}
  \caption {Comparison of our method with previous work across diverse object categories.}
    \label{fig4}
\end{figure}

We also analyze the per-category success rate of different zero-shot methods in \textbf{Figure \ref{fig4}}. RATE-Nav demonstrates superior performance across most goal categories, significantly outperforming other baseline methods. This consistent improvement across diverse object categories validates the generality and robustness of our approach in object search tasks.

\begin{table}[t]
\centering
\caption{Ablation study of different component of RATE-Nav on HM3D datset}
\renewcommand{\arraystretch}{1.2}
\begin{tabular}{cccccc}
\hline
\multirow{2}{*}{GPRS} & \multirow{2}{*}{REE}& \multirow{2}{*}{VP}  & \multicolumn{3}{c}{HM3D} \\
\cline{4-6}
 & & & SR $\uparrow$  & SPL $\uparrow $ & SSPL$\uparrow $ \\
\hline
 $\times$ & $\times$ & $\times$ & 45.3 & 20.2 & 25.1\\
\checkmark & $\times$ & $\times$ & 55.2 & 24.1 & 32.5\\
\checkmark & \checkmark & $\times$ & 57.7 & 26.7 & 33.2\\
\checkmark & $\times$ & \checkmark & 64.3 & 25.5 & 30.8\\
\hline
\checkmark & \checkmark & \checkmark & 67.8 & 31.3 & 38.6 \\
\hline
\end{tabular}

\label{table2}
\end{table}

\subsection{Ablation Study}
To comprehensively evaluate the effectiveness of our key modules, we conducted systematic ablation experiments in Table \ref{table2} on three core components: Geometric Predictive Region Segmentation (GPRS), Region-Based Exploration Estimation (REE), and VLM Perception (VP). Here, GPRS is responsible for semantics-based region division, REE handles explored area estimation, and VP provides visual-language model perception capabilities.
When all modules are removed, RATE-Nav degrades to a simple random exploration system. With the introduction of the GPRS module, the system can utilize region semantic maps to guide navigation. Without the REE module, the system relies solely on the occupancy map to calculate exploration coverage. Notably, the VP module not only provides conventional visual perception but also includes a crucial Re-perception mechanism - when the system believes it has found the target object, it performs a secondary confirmation through VLM to enhance target identification accuracy.

As shown in the Table \ref{tab: VLM}, we analyzed how the choice of Vision Language Models (VLM) and exploration rates affect navigation performance in the HM3D dataset. Results show that VLMs are crucial for macro-environment perception - Qwen-vl-max with optimal exploration rate significantly outperforms the baseline without VLM. The less capable Llama-vision shows performance degradation when the goal checking mechanism is removed. Exploration rate choice is equally important, as both low (0.5) and high (0.9) rates limit model performance. Additionally, removing the goal verification mechanism leads to significant performance drops in both models, highlighting the importance of goal detection.

Many existing approaches utilize semantic maps to evaluate the probability of candidate points. Our method innovatively introduces a region-based partitioning mechanism, which enables a more natural integration of object semantics into spatial features. Through comparative experiments on HM3D dataset, as shown in Table \ref{tab:scene}, we found that semantic maps incorporating region information demonstrate significant advantages over traditional methods: they can effectively differentiate between spatially adjacent areas belonging to different rooms, thereby greatly enhancing the system's understanding of the environment.

\begin{table}
    \centering
    \caption{The influence of vlm and exploration rate}
    \begin{tabular}{lccc}
        \hline
        \multirow{2}{*}{VLM} & \multirow{2}{*}{Rate}  & \multicolumn{2}{c}{HM3D}\\
        \cline{3-4}
        & & SR $\uparrow$ & SPL $\uparrow$ \\
        \hline
        w/o VLM & 0.7 &  35.1  & 14.7 \\
        LLama-vision & 0.7 & 60.1 & 26.2  \\
        Qwen-vl-max & 0.5 & 59.4 & 26.1 \\
        Qwen-vl-max & 0.7 & 67.8 & 31.3 \\
        Qwen-vl-max & 0.9 & 68.1 & 25.2 \\
        LLama w/o re-perception & 0.7 & 54.3 & 27.5 \\
        Qwen w/o re-perception & 0.7 & 60.3 & 34.2 \\
        \hline
    \end{tabular}
    
    \label{tab: VLM}
\end{table}

\begin{table}
    \centering
    \caption{The effect of region scene map on HM3D}
    \begin{tabular}{lll}
        \hline
        Method  & SR $\uparrow$ & SPL $\uparrow$ \\
        \hline
        w/o scene map   & 62.7 & 26.3 \\
        scene map w/o region info  &  65.3  & 30.1  \\
        region scene map   & 67.8   & 31.3 \\
        \hline
    \end{tabular}
    \label{tab:scene}
\end{table}

\subsection{VLM Perception Analysis}
\begin{figure}[h]
\centering
  \includegraphics[width=0.98\linewidth]{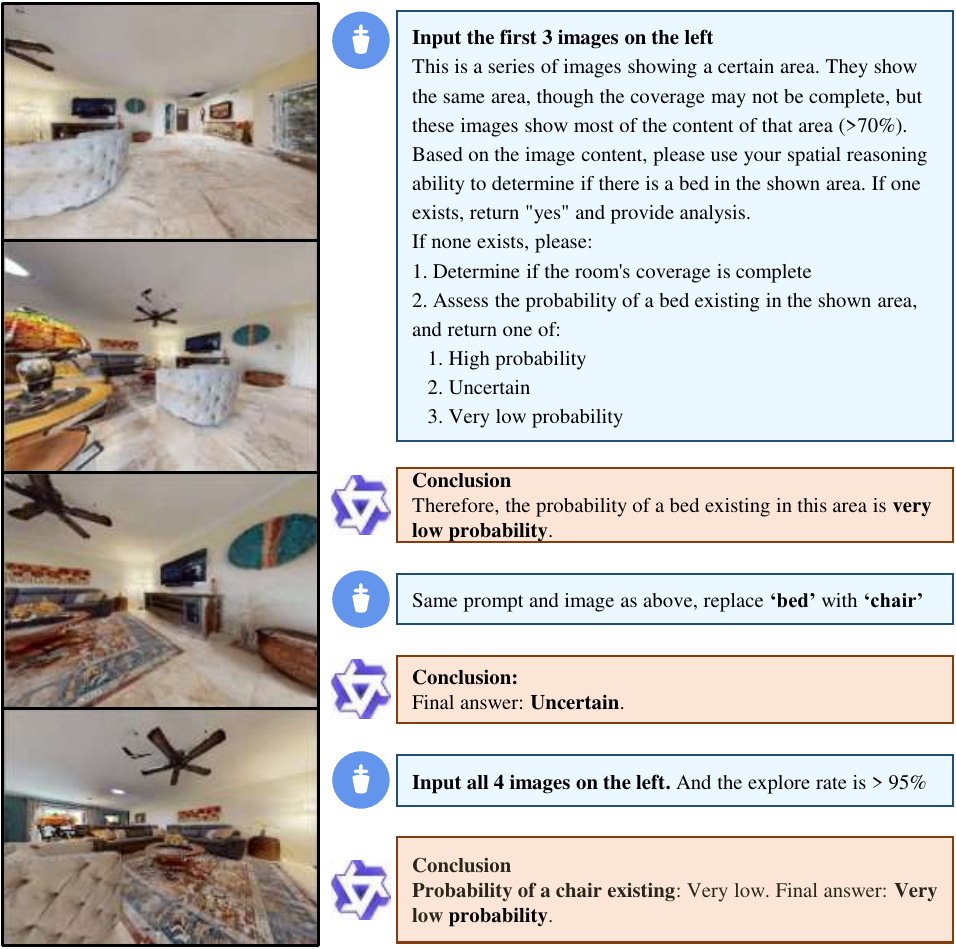}
  \caption{VLM Perception Analysis: Evaluated VLM performance across three input cases to test object recognition and classification capabilities.}
    \label{fig5}
\end{figure}

As shown in \textbf{Figure \ref{fig5}}, through a case study, we analyze the reasoning principles of Visual Language Models (VLM) in environmental perception. The case includes four perception images taken by the robot at different stages. For the target object "bed," due to its large semantic difference from the living room environment, the model can determine its absence using just the first three images. As for "chair," due to its higher probability of appearing in living rooms, the model needs further exploration to make a judgment. In fact, there is a table and chair set behind the viewpoint of the first image, validating the accuracy of VLM's reasoning. After obtaining more viewpoints and inputs, the model can finally determine that no chair exists in the currently visible area.

\section{Conclusion}
In this paper, we introduced RATE-Nav, a novel navigation method based on the law of diminishing returns. Our geometric predictive region segmentation method predicts unexplored areas using geometric features for region partitioning. Converting point-to-point to region-based processing with VLM spatial reasoning improves navigation accuracy and efficiency.
While our method effectively handles region-based references, future work should focus on enhancing spatial understanding for terms like "ahead" or "to the right" to enable more natural vision-language navigation.

\section{Limitations}
Although our method divides the occupancy map into distinct regions, enabling VLM outputs to correspond to specific map regions, this segmentation approach has limitations in incorporating VLM's more general and semantically rich descriptions. In our current work, VLM's spatial descriptions are confined to fixed, specific regions with associated exploration rates. However, in broader contexts, VLM's spatial descriptions tend to be more general, such as "forward" or "turn right." Future work could focus on localizing these linguistically described regions, which would allow us to extend our method to fields like vision-language navigation and image navigation, ultimately working toward a unified navigation framework. Additionally, while our method has only been tested on the Habitat simulator dataset, further exploration is needed to effectively adapt it to real-world applications.
\section*{Acknowledgments}
This work was sponsored by the Key Research and Development Program of Guangdong Province under grant No. 2021B0101400003, the National Key Research and Development Program of China under grant No.2023YFB4502701.

\bibliography{custom}

\appendix

\end{document}